\documentclass[lettersize,journal]{IEEEtran}
\usepackage{amsmath,amsfonts}
\usepackage{algorithmic}
\usepackage{array}
\usepackage[caption=false,font=normalsize,labelfont=sf,textfont=sf]{subfig}
\usepackage{textcomp}
\usepackage{stfloats}
\usepackage{url}
\usepackage{verbatim}
\usepackage{graphicx}
\usepackage{xcolor}  
\usepackage{graphicx}
\usepackage{colortbl}
\usepackage{threeparttable}
\usepackage{amsmath}
\usepackage[numbers,sort&compress]{natbib}
\usepackage{algorithm,algorithmic}
\usepackage{hyperref}
\definecolor{ccr}{RGB}{0,0,255}  
\hypersetup{hypertex=true,
    colorlinks=false,
    linkcolor=ccr,
    anchorcolor=ccr,
    citecolor=ccr}

\def\BibTeX{{\rm B\kern-.05em{\sc i\kern-.025em b}\kern-.08em
    T\kern-.1667em\lower.7ex\hbox{E}\kern-.125emX}}
\usepackage{balance}
\begin{document}
\title{LFMamba: Light Field Image Super-Resolution with State Space Model}
\author{Wang Xia,~\IEEEmembership{Student Member,~IEEE}, Yao Lu,~\IEEEmembership{Member,~IEEE}, Shunzhou Wang,~\IEEEmembership{Member,~IEEE}, Ziqi Wang, Peiqi Xia, and Tianfei Zhou
\thanks{This work was supported in part by the Special Projects in key areas of Guangdong Province (No.2022ZDZX1036), and in part by Shenzhen Peacock Plan. \textit{Corresponding author:} \textit{Yao Lu} (vis\_yl@smbu.edu.cn)}
\thanks{Wang Xia, Yao Lu, Ziqi Wang, Peiqi Xia, and Tianfei Zhou are with School of Computer, Beijing Institute of Technology, Beijing 100081, China (e-mail: wangxia@bit.edu.cn;
vis\_yl@sbmu.edu.cn; \{3220221026, 3120221009, tfzhou\}@bit.edu.cn).}
\thanks{Yao Lu and Wang Xia are also with Guangdong Laboratory of Machine Perception and Intelligent Computing, Department of Engineering, Shenzhen MSU-BIT University, Shenzhen 518172, China. }
\thanks{Shunzhou Wang is with School of Electronic and Computer Engineering, Shenzhen Graduate School, Peking University, Shenzhen 518055, China. (email: shunzhouwang@163.com). }
\thanks{Codes are public avilable at \url{https://github.com//stanley-313/LFMamba}.}
\thanks{Manuscript received xxx, 2024; revised xxx, 2024.}
}

\markboth{Journal of \LaTeX\ Class Files,~Vol.~**, No.~**, **~2024}%
{How to Use the IEEEtran \LaTeX \ Templates}

\maketitle

\begin{abstract}

Recent years have witnessed significant advancements in light field image super-resolution (LFSR) owing to the progress of modern neural networks. However, these methods often face challenges in capturing long-range dependencies (CNN-based) or encounter quadratic computational complexities (Transformer-based), which limit their performance. Recently, the State Space Model (SSM) with selective scanning mechanism (S6), exemplified by Mamba, has emerged as a superior alternative in various vision tasks compared to traditional CNN- and Transformer-based approaches, benefiting from its effective long-range sequence modeling capability and linear-time complexity. Therefore, integrating S6 into LFSR becomes compelling, especially considering the vast data volume of 4D light fields. However, the primary challenge lies in \emph{designing an appropriate scanning method for 4D light fields that effectively models light field features}. To tackle this, we employ SSMs on the informative 2D slices of 4D LFs to fully explore spatial contextual information, complementary angular information, and structure information. To achieve this, we carefully devise a basic SSM block characterized by an efficient SS2D mechanism that facilitates more effective and efficient feature learning on these 2D slices. Based on the above two designs, we further introduce an SSM-based network for LFSR termed LFMamba. Experimental results on LF benchmarks demonstrate the superior performance of LFMamba. Furthermore, extensive ablation studies are conducted to validate the efficacy and generalization ability of our proposed method. We expect that our LFMamba shed light on effective representation learning of LFs with state space models.
\end{abstract}


\section{Introduction}
\IEEEPARstart{L}{ight} field (LF) cameras record both the intensity and direction of light rays emitted from the world, enabling a wide range of valuable applications such as depth estimation \cite{depth}, refocusing \cite{refocus}, and salience detection \cite{saliency}. However, LF images captured by these cameras often sacrifice spatial resolution in individual sub-aperture images (SAIs) to capture multiple viewpoints simultaneously, limiting their usability in certain applications. Hence, there arises a necessity to enhance the spatial resolution of LF images, a task known as light field image super-resolution (LFSR). 

Taking advantage of CNNs, \citet{lLFCNN} made the first attempt to employ SRCNN \cite{srcnn} to achieve LFSR and light field image angular super-resolution (LFASR). Specifically, they first utilize SRCNN to enhance the resolution of each SAI, and then, adjacent HR SAIs are combined for novel view synthesis to increase the angular resolution by another SRCNN. Despite significant improvements over traditional methods \cite{project4, opt1, opt4, opt2, opt3}, super-resolving each SAI individually remains suboptimal, as it overlooks the rich angular information available across different SAIs. Subsequently, a series of more sophisticated networks have been proposed to fully exploit the abundant spatial-angular information inherent in light fields, further enhancing the reconstruction accuracy \cite{lfnet,reslf,LFSSR,ATO,megnet,dfnet,interNet,distg,iinet,sav,hlfsr}. Despite the progress they have achieved, the natural local reductive bias of convolutions limits the exploration of non-local relations of LFs. Additionally, the static learned kernel weights render these networks inflexible for out-of-distribution input light fields with various unseen scenarios.

Transformer \cite{transformer}, initially introduced in the field of natural language processing, has proven effective in modeling long-range dependencies and dynamically adjusting weights according to inputs. Consequently, it has found widespread application in various vision tasks, such as image classification \cite{vit,swin}, object detection \cite{dert,sun2021rethinking}, semantic segmentation \cite{sert,segmenter}, and image restoration \cite{liang2021swinir,zamir2022restormer}, becoming the de facto dominant architecture in vision areas. In recent years, a plethora of Tansformer-based methods have been proposed to tackle LFSR \cite{dpt,lft,wang2022local,wang2022multi,LF-DET, epit}. These methods commonly leverage the self-attention mechanism to establish long-range interrelationships from different views \cite{dpt, wang2022local, wang2022multi} or specific LF subspace domains \cite{lft, epit, LF-DET}, enabling effective excavation of global features and thereby further enhancing the reconstruction quality. However, the quadratic computational complexity of self-attention calculation hinders their efficient and comprehensive exploration of intrinsic properties in LFs. 

Recently, the State Space Models (SSMs) \cite{s4, smith2023simplified, gu2021combining, H3, mamba} have emerged as a novel and promising class of foundational architectures for sequence modeling. Among these, Mamba \cite{mamba}, a type of SSM endowed with selective mechanism (S6) that grants it content-based reasoning abilities, stands out by outperforming Transformer across various modalities including language, audio, and genomics in performance and efficiency. Inspired by these advancements, accompanying research has focused on exploring the potential of Mamba in vision tasks \cite{vim,liu2024vmamba,umamba,yang2024vivim,deng2024cu,guo2024mambair,shi2024vmambair,yang2024plainmamba}, demonstrating the feasibility and superiority of Mamba in vision fields. 

\begin{figure*}[ht]
    \centering
    \includegraphics[width=\linewidth]{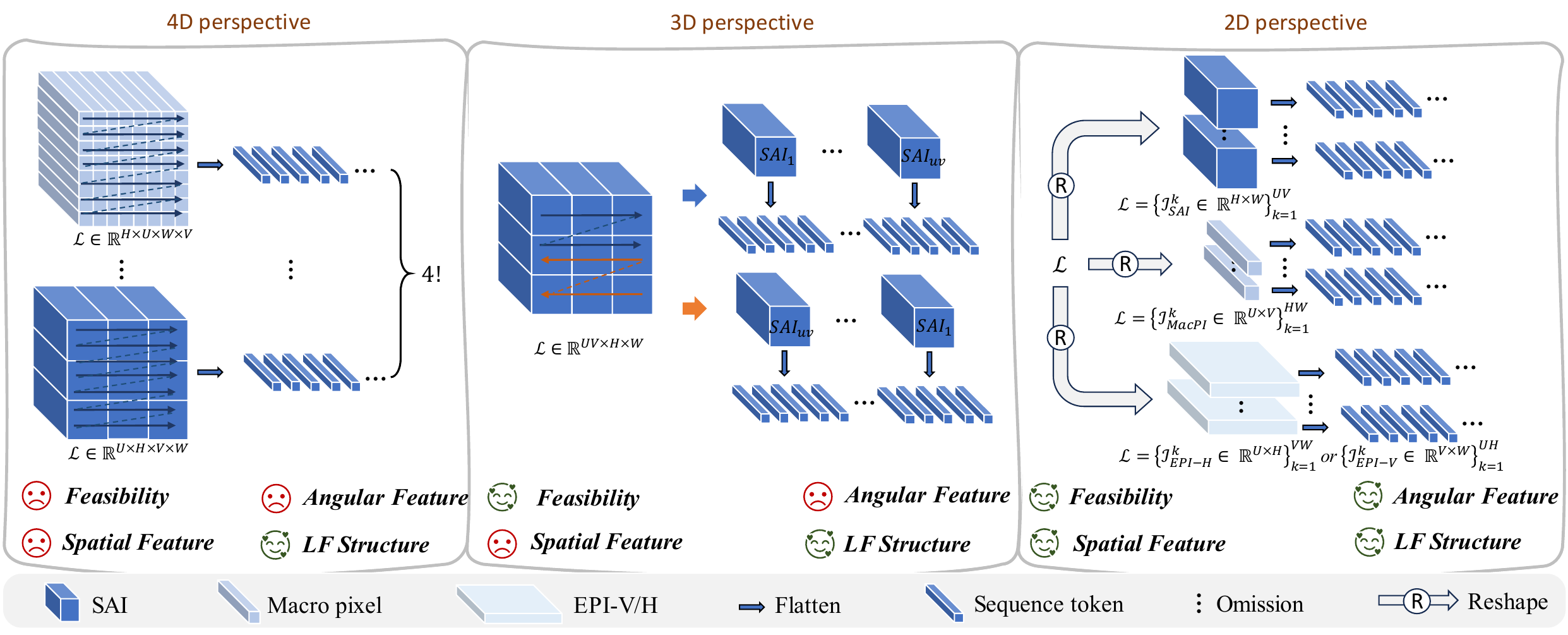}
    \caption{\textbf{Three different perspectives to model LFs using State Space Model.} The key point is how to flatten a 4D LF into a 1D sequence. \textbf{\emph{Left:}} Taking the 4D LF data as a whole and flatten it by different orders. \textbf{\emph{Mid:}} Taking LF as an 3D image sequence to explore the relationships between sub-aperture images. \textbf{\emph{Right:}} Taking LF as a combination of informative 2D data slices (i.e., sub-aperture image (SAI), macro-pixel image (MacPI), and epipolar plane image (EPI)) to fully capture spatial contextual information, complementary angular information, and structure information.}
    \label{fig:enter-label}
\end{figure*}

Naturally, a question arises: how can we integrate Mamba into LFSR effectively? To answer that, we should first recognize the prominent challenge is \emph{designing an appropriate way to scan 4D LFs that can fully explore the useful information of LFs}, considering the high dimensionality and complexity of LF data. To solve this problem, we initially devised three probable solutions from different perspectives as shown in Fig. \ref{fig:enter-label}. The first, and perhaps the most straightforward idea is to flatten the 4D light field into a 1D sequence by different orders. However, given a total of $4! = 24$ permutations, exhaustively searching the entire order space to ensure comprehensive modeling of the LFs is infeasible, and since the spatial and angular tokens are highly intertwined, the spatial contextual and angular information may be difficult to explore. The second viewpoint is to regard LFs as a 3D image sequence and flatten them by order as done in \cite{yang2024vivim} to explore the relationships of the image sequence. While this may uncover structural information considering the discrepancies between adjacent view images, it may also underutilize fundamental spatial contextual and angular information that is embedded in entire sequences. Lastly, drawing inspiration from visual state space models \cite{liu2024vmamba, guo2024mambair}, we can adopt Mamba to informative 2D slices (i.e., sub-aperture image (SAI), macro-pixel image (MacPI), and epipolar plane image (EPI)) of 4D LFs to independently extract spatial, angular, and EPI features, which is more effective and easier for implementation.

Based on the above analysis, we introduce LFMamba,  a novel SSM-based network for LFSR. Specifically, to achieve effective feature extraction on LFs' 2D slices, we first carefully devise a basic SSM block which is the core component of LFMamba. The basic SSM block integrates the proposed efficient SS2D, significantly reducing parameters with minimal performance drop. By utilizing the basic SSM block on LFs' 2D informative slices, LFMamba achieves a thorough exploration of diverse LF features. In summary, our contributions are as follows:
\begin{itemize}
    \item We integrate the State Space Model (SSM) into LFSR task for the first time by adopting SSM on LFs' 2D informative slices, which facilitates effective and comprehensive exploration of LFs' spatial contextual information, complementary angular information, and structure information.
    \item We design a basic SSM block featuring the proposed efficient SS2D mechanism. As the fundamental feature extractor of LF's 2D slices, it significantly reduces the parameters with little expressivity loss. 
    \item We propose an SSM-based network LFMamba for LFSR. Experimental results on LF benchmarks show LFMamba's efficacy and efficiency compared to state-of-the-art methods. We further demonstrate our method's generalization ability by applying LFMamba for LF angular SR task.
\end{itemize}

\section{Related Work}
\subsection{Light Field Image Super-Resolution}
Light field image super-resolution (LFSR), aiming at enhancing the spatial resolution of each SAI, has generated widespread attention in recent years. Since \citet{lLFCNN} first introduced CNNs into this realm, numerous learning-based methods have been proposed, showcasing progressive performance improvements. Inspired by LFCNN \cite{lLFCNN}, \citet{lf-dcnn} employed a more powerful single image super-resolution (SISR) network EDSR \cite{edsr} to increase the spatial resolution of SAIs, followed by another EPI-enhancement network to refine the results. After that, several multi-stream networks were proposed to exploit relationships between different SAIs. \citet{lfnet} designed two bidirectional recurrent networks to iteratively model spatial relations between horizontal and vertical SAIs. \citet{reslf, megnet} introduced a multi-branch network to learn the sub-pixel shift information of stacked SAIs in four directions. \citet{ATO} proposed an all-to-one approach implemented by a weight-shared multi-stream network, which combined all views for complementary information extraction and utilization. Apart from these methods, \citet{LFSSR} proposed spatial-angular separable convolution (SAS-Conv) to approximate 4D convolution while achieving efficiency. \citet{interNet} and \citet{iinet} proposed LF-InterNet and IINet, wherein both the spatial and angular features were extracted interactively. \citet{dfnet} utilized deformable convolution to align the spatial feature to the center view, addressing the disparity problem. More recently, \citet{sav} combined spatial-angular correlated convolution with SAS-Conv and proposed spatial-angular versatile convolution. \citet{distg} proposed a generic LF disentangling mechanism for LFSR, LF angular super-resolution, and depth estimation. \citet{hlfsr} further completed \cite{distg} by considering the inter-spatial and inter-angular relations.

Despite the intricate designs and architectures, performance reaches a bottleneck as these methods rely on convolution layers, which face crucial challenges in exploiting non-local spatial-angular information due to the limited receptive field capabilities. In response, some researchers have resorted to Transformer architectures, leveraging their robust long-range dependency modeling capabilities to break through this ceiling. \citet{dpt} proposed a two-branch network, which established long-range dependencies along horizontal and vertical SAI sequences. \citet{wang2022local} proposed a local-global aggregation network by combining CNNs and Transformer. Following the SAS-Conv scheme \cite{LFSSR}, \citet{lft} replaced convolutions with Transformer layers, resulting in superior performance. \citet{wang2022multi} employed Transformer to learn multi-granularity relationships between SAIs. \citet{epit} proposed EPIT which adopted Transformer to learn non-local spatial-angular correlation on EPI, achieving state-of-the-art performance. Despite the promising results, the quadratic computational complexity and memory footprint pose significant challenges for achieving efficient and effective LFSR. Notably, \citet{LF-DET} proposed a sub-sampled spatial Transformer inspired by PVT \cite{pvt} to reduce the computational overhead. However, the improvement is limited and the computational complexity remains quadratic. More advanced sub-quadratic complexity attention mechanisms (e.g., swin \cite{swin}, linear attention \cite{shen2021efficient}, flash attention \cite{dao2022flashattention}) or recent state space models (SSMs) \cite{s4, H3, mamba} with linear complexity, are desired in this field.

\subsection{State Space Model}
State Space Models (SSMs) \cite{s4, smith2023simplified, gu2021combining}, originating from classic control theory, have recently emerged as competitive backbones for state space transformation in deep learning. Their notable property of linearly scaling with sequence length in long-range dependency modeling has garnered significant interest from researchers. For instance, the structured state-space sequence model (S4) \cite{s4} is a pioneering work in deep state-space modeling for long-range dependency. Subsequently, the S5 layer \cite{smith2023simplified} was proposed, building upon S4 and introducing MIMO (Multiple Input Multiple Output) and efficient parallel scanning. Additionally, H3 \cite{H3} has achieved promising results, narrowing the performance gap between SSMs and Transformers in natural language processing. \citet{mehta2023long} integrate gated units into S4, further advancing the improvement of S4, and leading to the development of the gated state space layer. S4nd \cite{nguyen2022s4nd} is the pioneer in extending the applicability of SSMs beyond sequential data to continuous data domains such as images and videos, bridging the gap between sequential and spatial modeling.
More recently, Mamba \cite{mamba}, a data-dependent SSM with a selective mechanism (S6) and efficient hardware design, has surpassed Transformers in natural language processing tasks while demonstrating linear scaling with input length. Subsequently, numerous works have introduced Mamba into various vision tasks, including image classification \cite{vim, liu2024vmamba, yang2024plainmamba}, biomedical image segmentation \cite{yang2024vivim, umamba, wu2024ultralight}, and image restoration \cite{shi2024vmambair, guo2024mambair, deng2024cu}, and achieve comparable or superior performance in terms of accuracy and efficiency. 

\section{Method}
In this section, we start with an introduction to the state space model for light fields (Section. \ref{sec:III-A}). Then, we elaborate on the overall architecture of LFMamba in Section. \ref{sec:III-B}. After that, we expound on the process of how LFMamba learns spatial-angular features (Section. \ref{sec:III-C}) and structure features (Section. \ref{sec:III-D}) by our proposed basic SSM blocks. Then, we give a detailed introduction to the basic SSM block in Section. \ref{sec:III-E}. At last, in Sectio n. \ref{sec:III-G}, we summarize the overall algorithm for achieving LFSR using LFMamba.
\subsection{State Space Model for Light Field} \label{sec:III-A}
State Space Models (SSMs) are a type of linear time-invariant systems that map 1-D continuous simulation $x(t) \in \mathbb{R}^{N}$ into response $y(t) \in \mathbb{R}^{N}$, which the process can be formulated by linear ordinary differential equations (ODEs)
\begin{equation}
\begin{split}
    h^{'}(t) &= \mathbf{A}h(t) +\mathbf{B}x(t) ,\\
    y(t) &= \mathbf{C}h(t) ,
\end{split}
\label{eq:1}
\end{equation}
where the output $y(t)$ is derived from the input signal $x(t)$ and hidden state $h(t) \in \mathbb{R}^{N}$, and the parameters of the system include a state transition matrix $\mathbf{A} \in \mathbb{R}^{N \times N}$, and projection matrices $\mathbf{B} \in \mathbb{R}^{N}$ and $\mathbf{C} \in \mathbb{R}^{N}$. To integrate SSM into deep learning, a discretization process on ODEs is required, where in S4 \cite{s4}, $\mathbf{A}$ and $\mathbf{B}$ are discretized by zero-order hold (ZOH) to generate discrete parameters $\mathbf{\Bar{A}}$ and $\mathbf{\Bar{B}}$ using a time scale parameter $\Delta$
\begin{equation}
\begin{split}
    \mathbf{\Bar{A}} &= \exp{(\mathbf{\Delta}\mathbf{A})} , \\
    \mathbf{\Bar{B}} &= (\mathbf{\Delta} \mathbf{A})^{-1}(\exp{(\mathbf{\Delta} \mathbf{A})} - \mathbf{I}) \cdot \mathbf{\Delta} \mathbf{B} ,
\end{split}
\label{eq:2}
\end{equation}

After discretization, the discrete version of Eq. (\ref{eq:1}) can be rewritten as,
\begin{equation}
    \begin{split}
    h_{k} &= \mathbf{\Bar{A}}h_{k-1} + \mathbf{\Bar{B}}x_{k} , \\
    y_{k} &= \mathbf{C}h_{k} ,
\end{split}
\label{eq:3}
\end{equation}

At last, the output can be calculated in a global convolution form, enabling efficient parallelizable training.
\begin{equation}
    \begin{split}
    \mathbf{\Bar{K}} &= (\mathbf{C\Bar{B}}, \mathbf{C\Bar{A}\Bar{B}}, \cdots , \mathbf{C\Bar{A}^{L-1}\Bar{B}}) , \\
    y &= x*\mathbf{\Bar{K}}
\end{split}
\end{equation}
where $L$ is the length of the input sequence, and $\mathbf{\Bar{K}} \in \mathbb{R}^{L}$ denotes the structured convolutional kernel.

So far, the discretization of SSM facilitates efficient training while limited in content awareness compression due to its data-independent and time-invariant properties. Consequently, selective SSM (Mamba or S6) \cite{mamba} introduce data-dependent parameters $(\mathbf{B, C, \Delta})$ derived from the input data $x$ through a simple linear projection layer, ensuring its awareness of the contextual information embedded in the input data.

The 4D LF is parameterized by the two-plane model \cite{LF_render} as $\mathcal{L}(u,v,h,w) \in \mathbb{R}^{U\times V\times H \times W}$, where $U \times V$ represent the angular resolution, and $H \times W$ represent the spatial resolution. To apply SSM or Mamba to LFs, we regard 4D LF as combinations of four informative 2D slices $\mathcal{L} = \{\mathcal{L}_{SAI}, \mathcal{L}_{MacPI}, \mathcal{L}_{EPI-H}, \mathcal{L}_{EPI-V}\}$ as introduced in Fig. \ref{fig:enter-label}, where
\begin{equation}
    \begin{split}
        \mathcal{L}_{SAI}(:,:,h,w) &= \{\mathcal{I}_{SAI}^{i} \in \mathbb{R}^{H \times W}\}_{i=1}^{UV},\\
        \mathcal{L}_{MacPI}(u,v,:,:) &= \{\mathcal{I}_{MacPI}^{i} \in \mathbb{R}^{U \times V}\}_{i=1}^{HW},\\
        \mathcal{L}_{EPI-H}(u,:,h,:) &= \{\mathcal{I}_{EPI\_H}^{i} \in \mathbb{R}^{H \times U}\}_{i=1}^{WV},\\
        \mathcal{L}_{EPI-V}(:,v,:,w) &= \{\mathcal{I}_{EPI\_V}^{i} \in \mathbb{R}^{W \times V}\}_{i=1}^{HU}
    \end{split}
\end{equation}
Then, each 2D LF slice can be projected into the high-dimensional space and be flattened to generate a 1D sequence, which serves as the input for the SSM system. Taken $\mathcal{L}_{SAI}$ as an example, we first project it to produce a high-dimensional feature $\boldsymbol{L}_{SAI} = \{ \boldsymbol{I}_{SAI}^{i} \in \mathbb{R}^{H\times W\times C}\}_{i=1}^{UV}$. Each $\boldsymbol{I}_{SAI}^{i}$ is then reshaped into a 1D sequence of length $HW$ as $T_{SAI}^{i} = \{t_{k}^{i} \in \mathbb{R}^{C}\}_{k=1}^{HW}$. Subsequently, $t^{i}_{k}$ can be fed into an SSM system the same as $x_{k}$ in Eq. (\ref{eq:3}) to predict the corresponding output. In that way, the spatial contextual information of LFs is effectively extracted. Therefore, similar processes can be performed on $\mathcal{L}_{MacPI}$, $\mathcal{L}_{EPI-H}$, and $\mathcal{L}_{EPI-V}$ to fully exploit the angular information and structure information of LFs.

\begin{figure*}[!ht]
    \centering
    \includegraphics[width=0.9\linewidth]{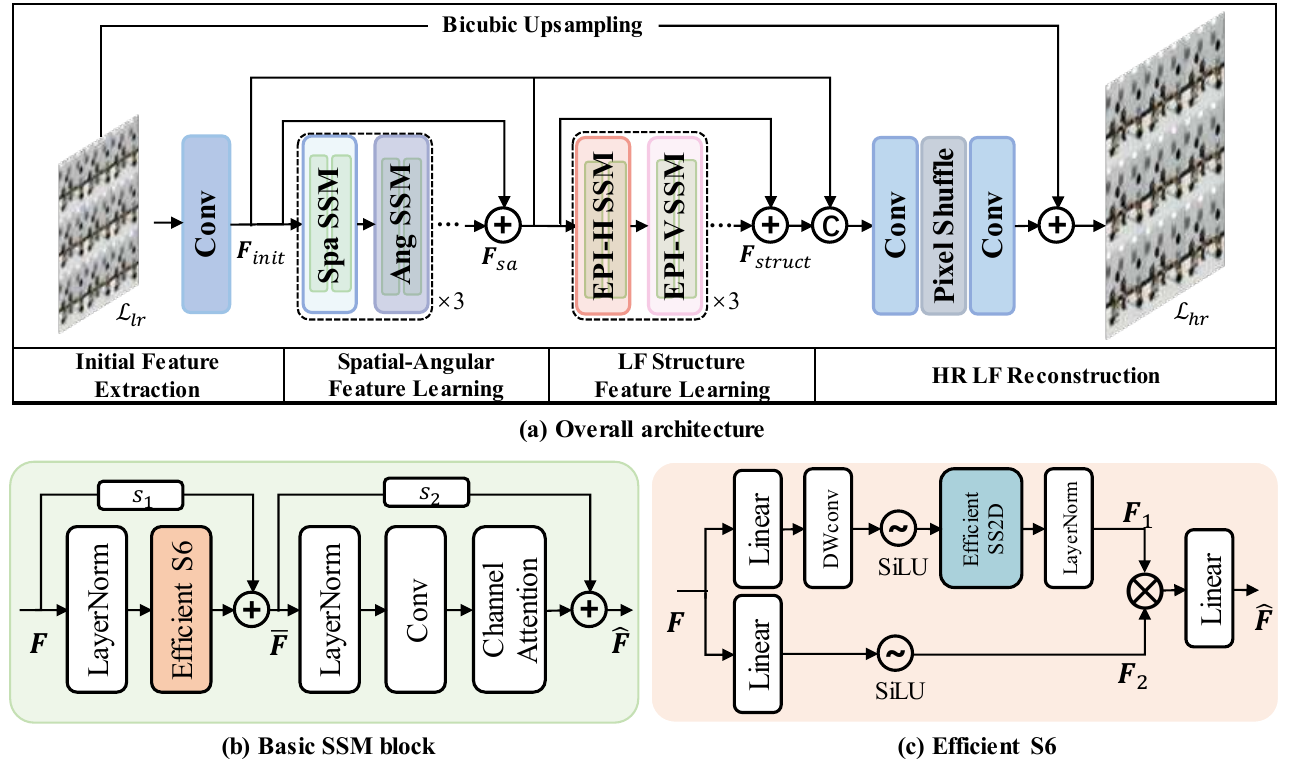}
    \caption{\textbf{LFMamba.} (a) The overall architecture of LFMamba. (b) The detailed structure of the core component, the basic SSM block. (c) The illustration of the proposed efficient S6.}
    \label{fig:net}
\end{figure*}

\begin{figure}[t]
    \centering
    \includegraphics[width=\linewidth]{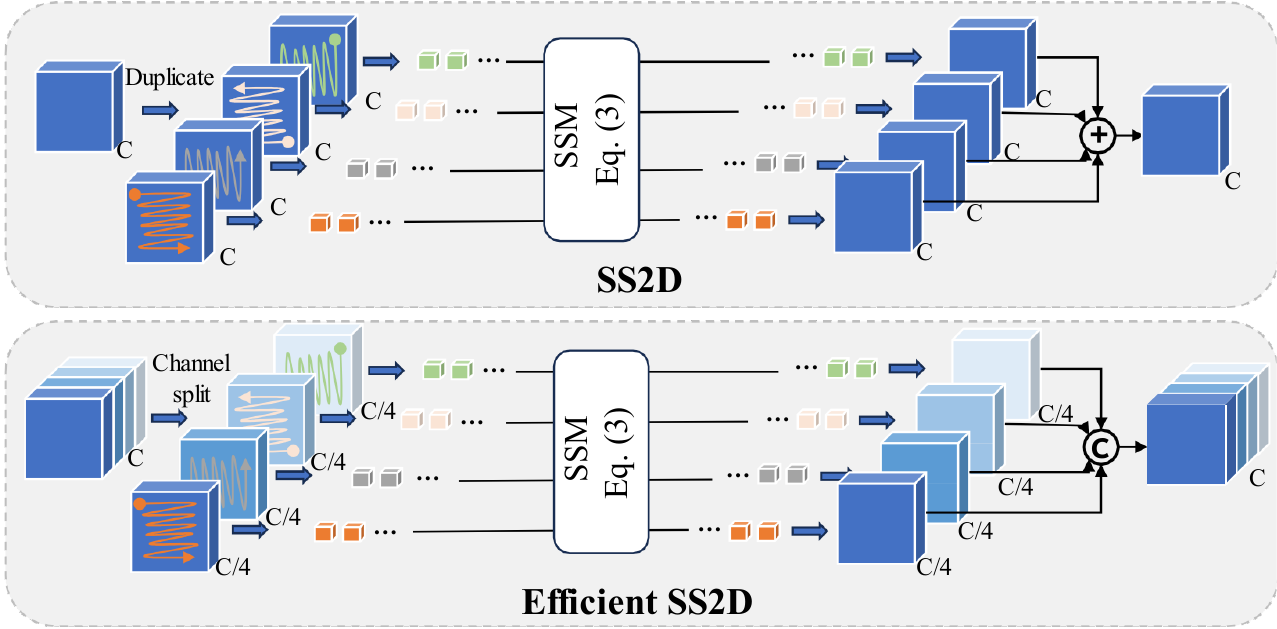}
    \caption{\textbf{Illustration of the original SS2D and our efficient SS2D.} \textbf{Up:} Original SS2D in visual state sapce model \cite{liu2024vmamba} copies the input four times for different scanning orders. \textbf{Down:} Our proposed efficient SS2D divides the input into four groups along the channel dimension for different scanning orders, which significantly reduces the parameters with little performance decline.}
    \label{fig:ss2d}
\end{figure}

Based on the above principle, we introduce a basic SSM block (see Section. \ref{sec:III-E}) that takes these 2D slices embedding as input. We further develop a novel SSM-based network based on this, termed LFMamba for LFSR, and we will present a detailed introduction of LFMamba in the following part.


\subsection{Overview of LFMamba} \label{sec:III-B}
For the problem of LFSR, given a low-resolution LF image $\mathcal{L}_{lr}\in\mathbb{R}^{U\times V\times H \times W}$, the goal of LFSR is to reconstruct the high-resolution LF image $\mathcal {L}_{hr}\in\mathbb{R}^{U\times V\times \alpha H \times \alpha W}$, where $\alpha$ is the upscaling factor. Following prior works \cite{distg, lft}, the LF image is organized as a $U\times V$ sub-aperture images array, and we convert it to YCbCr color space and only super-resolve the Y channel.

As shown in Fig. \ref{fig:net}(a), LFMamba mainly comprises four parts: Initial Feature Extraction module (IFE), Spatial-Angular Feature Learning module (SAFL), LF Structure Feature Learning module (LSFL), and HR LF Reconstruction module (HLFR). Specifically, following \cite{lft}, the input LF images $\mathcal{L}_{lr}$ are first processed by cascaded convolution layers to extract initial spatial features $\boldsymbol{F}_{init} \in \mathbb{R}^{U\times V\times H \times W \times C}$, where $C$ denotes the channel dimension and is set to 64 in our implementation. Then, the features will be further transferred to the SAFL module to capture the spatial contextual feature and complementary angular information. After that, the middle features $\boldsymbol{F}_{sa} \in \mathbb{R}^{U\times V\times H \times W \times C}$ will processed by the LSFL module to incorporate the structural information of LFs to generate $\boldsymbol{F}_{struct} \in \mathbb{R}^{U\times V\times H \times W \times C}$. Subsequently,  we fuse the above three features via concatenation followed by a $1\times 1$ convolution layer to achieve multi-level feature utilization.
 \begin{equation} \label{eq:6}
     \boldsymbol{F}_{fuse} = {\rm{Conv}}({\rm{Concat}}(\boldsymbol{F}_{init}, \boldsymbol{F}_{sa}, \boldsymbol{F}_{struct}))
 \end{equation}
 Lastly, following most precious works \cite{lft, epit, distg}, the fused feature $\boldsymbol{F}_{fuse}$ will undergo a pixel shuffling layer and a convolution layer to generate the final HR LF image $\mathcal {L}_{hr}$.

 \subsection{Spatial-Angular Feature Learning} \label{sec:III-C}
 The SAFL module endeavors to comprehensively integrate the spatial contextual information and complementary angular information inherent in LFs. To this end, we devise a spatial SSM block and an angular SSM block, and employ a simple but effective spatial-angular separable modeling scheme \cite{LFSSR, lft}, in which case, as illustrated in Fig. \ref{fig:net}(a), the spatial SSM block and the angular SSM block are adopted alternately for three times with local skip connection. 
 \begin{equation}
 \begin{split}
    \boldsymbol{F}_{sa} &= \rm{H_{SAFL}}(\boldsymbol{F}_{init}) + \boldsymbol{F}_{init} , \\
    \rm{H_{SAFL}}(\cdot) &= \rm{H_{spa}^{3}(H_{ang}^{3}}( \cdots H_{spa}^{1}(H_{ang}^{1}(\cdot)))) 
    \end{split}
 \end{equation}
 where $\rm{H_{SAFL}}(\cdot)$ denotes the SAFL module, $\rm{H_{spa}}(\cdot)$ and $\rm{H_{ang}}(\cdot)$ represent the spatial SSM block and angular SSM block, respectively.
 
  The spatial SSM block is designed for spatial contextual information extraction on each individual SAI. Specifically, for a given input 4D LF feature $\boldsymbol{F} \in \mathbb{R}^{U\times V\times H \times W \times C}$, we first reshape it to its 2D SAI form $\boldsymbol{F}_{SAI} \in \mathbb{R}^{UV\times H \times W \times C}$ to facilitate more effective spatial feature extraction, where $UV$ denotes the batch size. After that, two basic SSM blocks (see Fig. \ref{fig:net}(b) and detail introduction in Section. \ref{basic SSM}) are adopted for spatial feature extraction. After that, the feature is reshaped back to its 4D form for further processing, which can be formulated as
\begin{equation}
\begin{split}
     \boldsymbol{F}_{SAI} &= \rm{Reshape}(\boldsymbol{F}), \\
     \boldsymbol{\hat{F}}_{SAI} &= \rm{H_{SSM}^{2}(H_{SSM}^{1}}(\boldsymbol{F}_{SAI})), \\
     \boldsymbol{\hat{F}} &= \rm{Reshape}(\boldsymbol{\hat{F}}_{SAI})
\end{split}
\end{equation}
 where $\boldsymbol{\hat{F}}_{SAI}$ and $\boldsymbol{\hat{F}}$ denotes the enhanced features, and $\rm{H_{SSM}}$ denote the basic SSM block.
 
The angular SSM block adopts a similar processing approach, we reshape the input 4D LF feature to its 2D macro pixel form (i.e., $\mathbb{R}^{HW\times U \times V \times C}$) to achieve angular feature learning by another two basic SSM blocks and finally reshaped the feature back to its 4D representation.

\subsection{LF Structure Feature Learning}\label{sec:III-D}

 After extracting spatial contextual and complementary angular information of LFs, further excavating the structural property of LFs is crucial for reconstructing accurate HR LF images. In the LSFL module, we empirically adopt basic SSM blocks on horizontal and vertical EPIs as they are highly correlated with the scene depth, to exploit the structure information of LFs. As depicted in Fig. \ref{fig:net}(a), in the LFSL module, we alternately perform EPI-H and EPI-V SSM block three times to achieve LF structure learning. The EPI-H/V SSM block also follows the previous process to reshape the input LF feature to the corresponding forms (i.e., $\boldsymbol{F}_{EPI\_H} \in \mathbb{R}^{VW \times U \times H \times C}, \boldsymbol{F}_{EPI\_V} \in \mathbb{R}^{UH \times V \times W \times C}$) before being fed into the basic SSM blocks. Similar to \cite{epit}, we share the parameters between horizontal and vertical EPI SSM blocks. Then, we will give a detailed introduction to the core component of LFMamba, the basic SSM block.

\subsection{Basic SSM Block}\label{basic SSM} \label{sec:III-E}
 As the fundamental component of LFMamba, the basic SSM block is responsible for effectively establishing long-range dependencies of each 2D LF slice as an alternative to Transformer block. As illustrated in Fig. \ref{fig:net}(b). Following the design philosophy of Transformer, a two-stage structure is adopted in the basic SSM block. At the first stage, given an input 2D feature $\boldsymbol{F} \in \mathbb{R}^{H \times W \times C}$ (batch dimension is omitted), we employ a LayerNorm (LN) operation followed by our proposed efficient S6 block to establish long-range dependencies of each pixel, and we adopt learnable skip connection with scale factor $s_{1} \in \mathbb{R}^{C}$ between the input and the output 
 \begin{equation}
     \begin{split}
         \Bar{\boldsymbol{F}} = {\rm{S6}(LN}(\boldsymbol{F})) + s_{1} \cdot \boldsymbol{F} ,
     \end{split}
 \end{equation}
 
 Then, in the second stage, after normalizing $\Bar{\boldsymbol{F}}$ by LayerNorm, following \cite{guo2024mambair}, we employ a convolution (Conv) layer to capture local details followed by a channel attention (CA) layer to enhance the channel interaction. At last, we also utilize a learnable scale factor $s_{2}$ on residual learning in this stage.
  \begin{equation}
     \begin{split}
         \Hat{\boldsymbol{F}} = {\rm{CA}(Conv(LN}(\Bar{\boldsymbol{F}})) + s_{2} \cdot \Bar{\boldsymbol{F}}
     \end{split}
 \end{equation}

\textbf{Efficient S6.} The vanilla Mamba was initially introduced for casual modeling of 1D sequences, which poses significant challenges when adapting to non-causal data such as images \cite{liu2024vmamba}. To solve this problem, plenty of vision mambas have proposed various multi-direction 2D scanning strategies to alleviate directional sensitiveness, such as bidirectional scan \cite{vim}, cross-scan \cite{liu2024vmamba}, continuous-scan \cite{yang2024plainmamba}, and etc. Inspired by \cite{liu2024vmamba}, we introduce the efficient S6 block for efficient and effective handling of 2D LF informative slices

 As illustrated in Fig. \ref{fig:net}(c), the efficient S6 block follows the design paradigm of the Visual State-Space (VSS) block introduced in \cite{liu2024vmamba} which employs a two-stream structure.
 Despite the strong modeling capability presented by \cite{liu2024vmamba}, we surprisingly find that directly utilizing the original SS2D in VSS block \cite{liu2024vmamba} achieves sub-optimal results compared to existing leading methods with more parameters. Improving the performance by simply adding more layers or expanding the channel dimension will result in a larger model size and longer execution time which is unsatisfactory. To this end, we propose a novel efficient SS2D (ESS2D) mechanism that is integrated into the efficient S6 block. Therefore, given an input feature $\boldsymbol{F} \in \mathbb{R}^{H \times W \times C}$, the whole process of efficient S6 can be described as
   \begin{equation}
     \begin{split}
         \boldsymbol{F}_{1} &= {\rm{LN}(ESS2D(SiLU(DWConv(Linear(}{\boldsymbol{F}}))))) , \\
         \boldsymbol{F}_{2} &= \rm{SiLU(Linear(}\boldsymbol{F})) , \\
         \hat{\boldsymbol{F}} &= \rm{Linear}(\boldsymbol{F}_{1} \odot \boldsymbol{F}_{2})
     \end{split}
 \end{equation}
 where $\rm{SiLU}$ denotes the sigmoid-weighted linear unit activation function \cite{silu}, $\rm{DWConv}$ denotes the depth-wise convolution, and $\odot$ represents the Hadamard product.
 
 As illustrated in Fig. \ref{fig:ss2d}, the upper subplot shows the original SS2D proposed in \cite{liu2024vmamba}. To facilitate modeling non-casual 2D images, SS2D duplicates the input feature four times, and each copy will be flattened into a 1D sequence in different orders for long-term dependency learning. Finally, the output can be obtained by adding them together. In contrast, ESS2D divides the input data into four groups along the channel dimension, then each data group is flattened into a 1D sequence in four directions for further feature extraction. Eventually, each data sequence is restored to the original 2D data and concatenate back together. This simple alternation brings considerable declines of parameters, which mainly occurs on the two linear projection layers in Fig. \ref{fig:net}(c) while sacrificing little learning ability. This allows us to ultimately cascade two basic SSM blocks in Spatial/Angular/EPI-H/V SSM blocks to enhance the modeling capability further.

 Notably, our proposed ESS2D is similar to the Parallel Vision Mamba Layer (PVM) in \cite{wu2024ultralight} but has some major differences. At first, the PVM also divides the input data into four groups along channel dimension, but the scan order for these four parts is the same. Second, the PVM is a multi-branch structure equipped with four parallel Mamba blocks, the Mamba block corresponds to our proposed efficient S6 block, which means that the division operation of the input in \cite{wu2024ultralight} happens before feeding into the Mamba block, while ours happens inside the efficient S6 block. Please refer to \cite{wu2024ultralight} for more detailed introduction of PVM.

\begin{table*}[h]
    \centering
    \caption{\textbf{Quantitative comparison (PSNR / SSIM) of different methods for $\times2$ and $\times4$ SR.} The best and the second best results are marked with {\color{red}red} and {\color{blue}blue} colors. The parameters (\#Param.) and FLOPs are calculated on an input light field with size $5\times5\times32\times32$.}
    \resizebox{1\textwidth}{!}{
    \begin{tabular}{|l|c|c|c|cccccc|}
        \hline
       Method & Scale & \#Param.(M) & FLOPs(G) & EPFL&  HCInew &  HCIold & INRIA & STFgantry & Average \\
         \hline
         Bicubic & $\times 2$ & - & - & 29.50/.9350 & 31.69/.9335 & 37.46/.9776 & 31.10/.9563 & 30.82/.9473 & 31.11/.9542\\
         RCAN \cite{rcan} & $\times 2$ & 15.3 & 389.75 & 33.16/.9635 & 34.98/.9602 & 41.05/.9875 & 35.01/.9769 & 36.33/.9825 & 36.11/.9742\\
         \hline
         resLF \cite{reslf} & $\times 2$ & 7.98 & 37.06 & 32.75/.9672 & 36.07/.9715 & 42.61/.9922 & 34.57/.9784 & 36.89/.9873 & 36.58/.9793\\
         LFSSR \cite{LFSSR} & $\times 2$ & 0.88 & 25.70 &33.69/.9748 & 36.86/.9753 & 43.75/.9939 & 35.27/.9834 & 38.07/.9902 & 37.73/.9835\\
         LF-InterNet \cite{interNet} & $\times 2$ & 5.04 & 47.46 &34.14/.9761 & 37.28/.9769 & 44.45/.9945 & 35.80/.9846 & 38.72/.9916 & 38.08/.9847 \\
         LF-ATO \cite{ATO} & $\times 2$ & 1.22 & 597.66 & 34.27/.9757 & 37.24/.9767 & 44.20/.9942 & 36.15/.9842 & 39.64/.9929 & 38.15/.9843 \\
         MEG-Net \cite{megnet} & $\times 2$ & 1.69 & 48.40 & 34.34/.9773 & 37.42/.9777 & 44.08/.9942 & 36.09/.9849 & 38.77/.9915 & 38.14/.9851 \\
         LF-DFNet \cite{dfnet} & $\times 2$ & 3.94 & 57.22 & 34.44/.9766 & 37.44/.9786 & 44.23/.9943 & 36.36/.9841 & 39.61/.9935 & 38.41/.9854 \\
         IINet \cite{iinet} & $\times 2$ & 4.84 & 56.16 & 34.68/.9773 & 37.74/.9790 & 44.84/.9948 & 36.57/.9853 & 39.86/.9936 & 38.74/.9857 \\
         LF-SAV \cite{sav} & $\times 2$ & 1.22 & 34.65 & 34.62/.9772 & 37.43/.9776 & 44.22/.9942 & 36.36/.9849 & 38.69/.9914 & 38.26/.9851 \\
         DistgSSR \cite{distg} & $\times 2$ & 3.53 & 64.11 & 34.81/.9787 & 37.96/.9796 & 44.94/.9949 & 36.59/.9859 & 40.40/.9942 & 38.94/.9866 \\
         HLFSR \cite{hlfsr} & $\times 2$ & 13.72 & 167.40 & 35.31/.9800 & 38.32/.9807 & 44.98/.9950 & 37.06/.9867 & 40.85/.9947 & 39.30/.9874\\
         \hline
         DPT \cite{dpt} & $\times 2$ & 3.73 & 65.34 & 34.48/.9758 & 37.35/.9771 & 44.31/.9943 & 36.40/.9843 & 39.52/.9926 & 38.40/.9848 \\
         LFT  \cite{lft} & $\times 2$ & 1.11 & 56.16 & 34.80/.9781 & 37.84/.9791 & 44.52/.9945 & 36.59/.9855 & 40.51/.9941 & 38.85/.9863 \\
         EPIT \cite{epit} & $\times 2$ & 1.42 & 69.71 & 34.83/.9775 & 38.23/{\color{blue}.9810} & {\color{blue}45.08}/.9949 & 36.67/.9853 & {\color{red}42.17/.9957} & 39.40/.9877\\
         LF-DET \cite{LF-DET} & $\times 2$ & 1.59 & 48.50 & 35.26/.9797 & 38.31/.9807 & 44.99/{\color{blue}.9950} & 36.95/.9864 & {\color{blue}41.76/.9955} & {\color{blue}39.45}/.9875\\
         
         \rowcolor{gray!30}LFMamba & $\times 2$ & 2.15 & 62.95 & {\color{blue}35.75/.9824} & {\color{blue}38.36/.9810} & 44.98/{\color{blue}.9950} & {\color{blue}37.07/.9876} & 40.95/.9948 & 39.42/{\color{blue}.9882}\\
         \rowcolor{gray!30}LFMamba {\dag} & $\times 2$ & 2.15 & 62.95 & {\color{red}35.84/.9832} & {\color{red}38.59/.9816} & {\color{red}45.20/.9952}& {\color{red}37.19/.9880} & 41.15/.9950&{\color{red}39.59/.9886}\\
         \hline
         Bicubic & $\times 4$ & - & - & 25.14/.8311 & 27.61/.8507 & 32.42/.9335 & 26.82/.8860 & 25.93/.8431 & 27.58/.8701\\
         RCAN \cite{rcan} & $\times 4$ & 15.4 & 391.25 & 27.88/.8863 & 29.63/.8886 & 35.20/.9548 & 29.76/.9276 & 28.90/.9131 & 30.27/.9141\\
         \hline
         resLF \cite{reslf} & $\times 4$ & 8.64 & 39.70 &28.27/.9035 & 30.73/.9107 & 36.71/.9682 & 30.34/.9412 & 30.19/.9372 & 31.25/.9322\\
         LFSSR \cite{LFSSR} & $\times 4$ & 1.77 & 128.44 &28.27/.9118 & 30.72/.9145 & 36.70/.9696 & 30.31/.9467 & 30.15/.9426 & 31.23/.9370\\
         LF-InterNet \cite{interNet} & $\times 4$ & 5.48 & 50.10 & 28.67/.9162 & 30.98/.9161 & 37.11/.9716 & 30.64/.9491 & 30.53/.9409 & 31.58/.9388\\
         LF-ATO \cite{ATO} & $\times 4$ & 1.36 & 686.99 &28.52/.9115 & 30.88/.9135 & 37.00/.9699 & 30.71/.9484 & 30.61/.9430 &  31.54/.9373\\
         MEG-Net \cite{megnet} & $\times 4$ & 1.77 & 102.20 &28.74/.9160 & 31.10/.9177 & 37.28/.9716 & 30.66/.9490 & 30.77/.9453 & 31.71/.9399\\
         LF-DFNet \cite{dfnet} & $\times 4$ & 3.99 & 57.31 &28.77/.9165 & 31.23/.9196 & 37.32/.9718 & 30.83/.9503 & 31.15/.9494 & 31.86/.9415\\
         IINet \cite{iinet} & $\times 4$ & 4.88 & 57.42 &29.11/.9188 & 31.36/.9208 & 37.62/.9734 & 31.08/.9515 & 31.21/.9502 & 32.08/.9429\\
         LF-SAV \cite{sav} & $\times 4$ & 1.54 & 115.80 &29.37/.9223 & 31.45/.9217 & 37.50/.9721 & 31.27/.9531 & 31.36/.9505 & 32.19/.9439\\
         DistgSSR \cite{distg} & $\times 4$ & 3.58 & 65.41 &28.99/.9195 & 31.38/.9217 & 37.56/.9732 & 30.99/.9519 & 31.65/.9535 & 32.11/.9440\\
         HLFSR \cite{hlfsr} & $\times 4$ & 13.87 & 182.52 &29.20/.9222 & 31.57/.9238 & 37.78/.9742 & 31.24/.9534 & 31.64/.9537 & 32.29/.9455\\
         \hline
         DPT \cite{dpt} & $\times 4$ & 3.78 & 66.55 &28.93/.9170 & 31.19/.9188 & 37.39/.9721 & 30.96/.9503 & 31.14/.9488 & 31.92/.9414\\
         LFT \cite{lft} & $\times 4$ & 1.16 & 57.60 &29.25/.9210 & 31.46/.9218 & 37.63/.9735 & 31.20/.9524 & 31.86/.9548 & 32.28/.9447\\
         EPIT \cite{epit} & $\times 4$ & 1.47 & 74.96 &29.34/.9197 & 31.51/.9231 & 37.68/.9737 & 31.37/.9526 & {\color{red}32.18}/{\color{blue}.9571} & 32.40/.9452\\
         LF-DET \cite{LF-DET} & $\times 4$ & 1.69 & 51.20 &{29.47/.9230} & {31.56/.9235} & {37.84/.9744} & {31.39/.9534} & {\color{blue}32.14}/{\color{red}.9573} & {32.48/.9463}\\
         
         \rowcolor{gray!30}LFMamba & $\times 4$ & 2.30 & 66.90 & {\color{blue}29.84/.9256} & {\color{blue}31.70/.9249} & {\color{blue}37.91/.9748}& {\color{blue}31.81/.9551} & {31.85/.9554} & {\color{blue}32.62/.9472}\\

         \rowcolor{gray!30}LFMamba {\dag} & $\times 4$ & 2.30 & 66.90 & {\color{red}29.95/.9275} & {\color{red}31.86/.9265} & {\color{red}38.08/.9755}& {\color{red}31.90/.9563} & 32.04/.9568 & {\color{red}32.77/.9485}\\
         \hline
    \end{tabular}
    }
    \begin{tablenotes}
        \footnotesize
        \item Notes: $\dag$ denotes geometry assembling strategy.
    \end{tablenotes}
    \label{tab:sota}
\end{table*}

\subsection{Algorithm of LFMamba for LFSR} \label{sec:III-G}
In summary, the detailed algorithm for achieving super-resolution of light field images using LFMamba is shown in Algorithm 1.
\begin{algorithm}
    \caption{LFMamba for LFSR}
    \renewcommand{\algorithmicrequire}{\textbf{Input:}}
    \renewcommand{\algorithmicensure}{\textbf{Output:}}
    \label{alg:lfsr}
    \begin{algorithmic}[1]
        \REQUIRE Low resolution light field image $\mathcal{L}_{lr}$
	\ENSURE High resolution light field image $\mathcal{L}_{hr}$

        \STATE {\color{gray!80!black}{\# Step 1: Project $\mathcal{L}_{lr}$ to embedding space}} 
        \STATE \hspace{1em} $\boldsymbol{F}_{init} = H_{IFE}(\mathcal{L}_{lr})$ {\color{gray!80!black}\COMMENT{IFE module}}
        
        \STATE {\color{gray!80!black}{\# Step 2: Extract spatial and angular information}}
        \STATE \hspace{1em} Let $\boldsymbol{F}^{0} = \boldsymbol{F}_{init}$
        \STATE \hspace{1em} \textbf{for} {$i=1,2,3$} \textbf{do}
        \STATE \hspace{2em} $\boldsymbol{F}^{i} =  H_{spa}^{i}(H_{ang}^{i}(\boldsymbol{F}^{i-1}))$ {\color{gray!80!black}\COMMENT{SAFL module}}
        \STATE \hspace{1em} \textbf{end for}
        \STATE \hspace{1em} $\boldsymbol{F}_{sa} = \boldsymbol{F}_{init} + \boldsymbol{F}^{2}$
        
        \STATE {\color{gray!80!black}{\# Step 3: {Extract structure information}}}
        \STATE \hspace{1em} Let $\boldsymbol{F}^{0} = \boldsymbol{F}_{sa}$
        \STATE \hspace{1em} \textbf{for} {$i=1,2,3$} \textbf{do}
        \STATE \hspace{2em} $\boldsymbol{F}^{i} =  H_{epi-h}^{i}(H_{epi-v}^{i}(\boldsymbol{F}^{i-1}))$  {\color{gray!80!black}\COMMENT{LSFL module}}
        \STATE \hspace{1em} \textbf{end for}
        \STATE \hspace{1em} $\boldsymbol{F}_{struct} = \boldsymbol{F}_{sa} + \boldsymbol{F}^{2}$
        
        \STATE {\color{gray!80!black}{\# Step 4: Generate $\boldsymbol{F}_{fuse}$ using Eq. (\ref{eq:6})}} 
        \STATE \hspace{1em} $\boldsymbol{F}_{fuse} = H_{fuse}(\boldsymbol{F}_{init}, \boldsymbol{F}_{sa}, \boldsymbol{F}_{struct})$ 

        \STATE {\color{gray!80!black}{\# Step 5: Generate $\mathcal{L}_{hr}$}} {\color{gray!80!black}\COMMENT{HLFR module}}
        \STATE \hspace{1em} $\mathcal{L}_{hr} = H_{HLFR}(\boldsymbol{F}_{fuse})$ 
        
    \end{algorithmic}
\end{algorithm}

\section{Experiment}
\subsection{Dataset and Implementation Detials}
Following prior works \cite{distg, dfnet}, we select five LFSR benchmarks (i.e., EPFL \cite{epfl}, HCIold \cite{hciold}, HCInew \cite{hcinew}, INRIA \cite{inria}, STFgantry \cite{STFgantry}) for training and evaluation. More specifically, we use the central $5 \times 5$ sub-aperture images from the original LFs and crop them into $32 \times 32$ and $64 \times 64$ patches for $2\times$ and $4\times$ SR, followed by bicubic downsampling to generate low-resolution $16 \times 16$ patches. Data augmentation is conducted by random 90-degree rotation and flipping in horizontal and vertical directions. We select $L_{1}$ loss as our loss function, and Adam as our optimizer with $\beta_1 =  0.9$ and $\beta_2 = 0.999$. The initial learning rate is set to $2\times10^{-4}$ and will be halved every 15 epochs in total 60 epochs. The batch size is set to 2 for training. All the experiments are conducted on one PC with an RTX 2080 Ti GPU.

\begin{figure*}[t]
    \centering
    \includegraphics[width=0.9\linewidth]{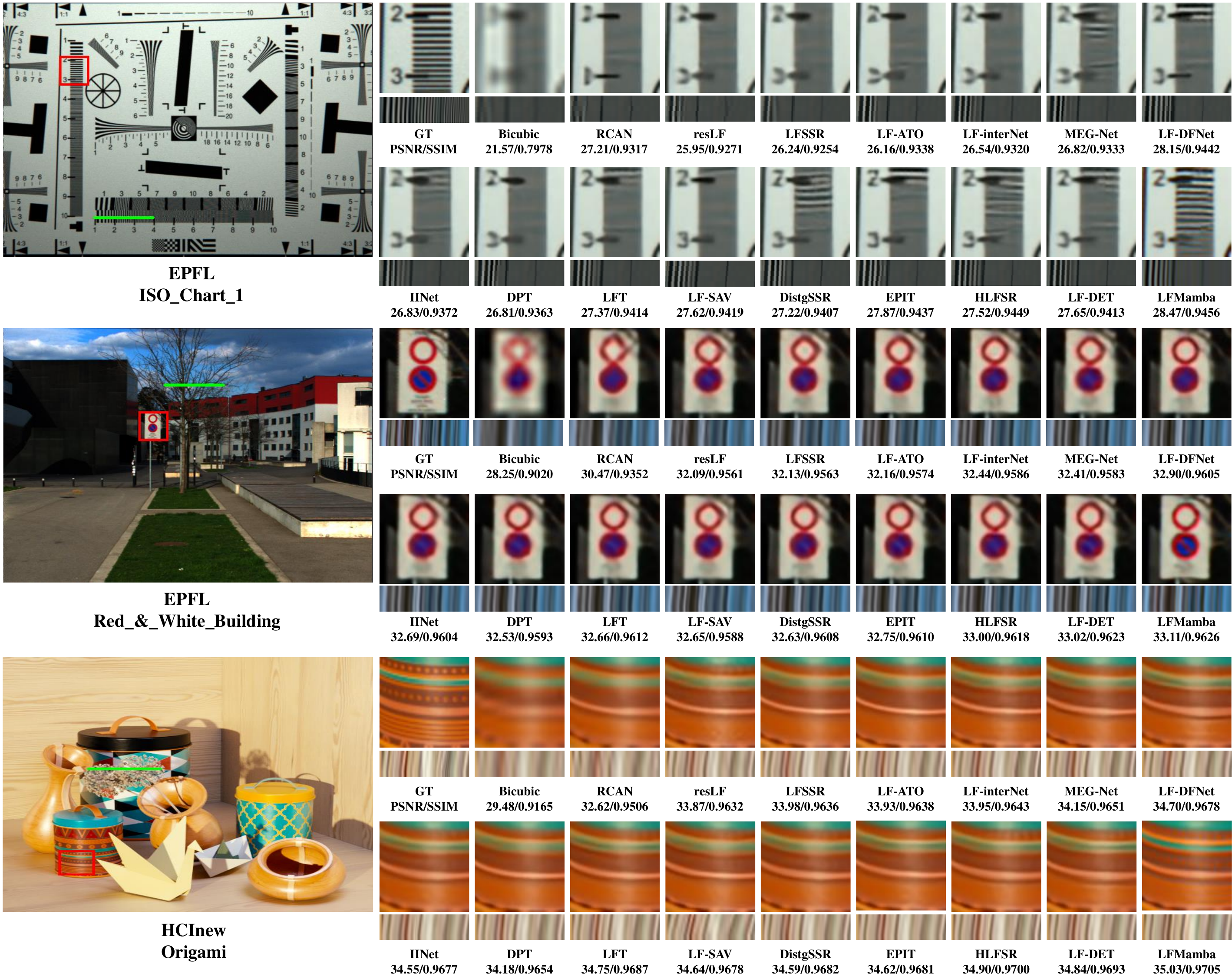}
    \caption{\textbf{Visual comparisons of different methods for $4 \times$ LFSR.} The first column shows the HR central view image and the rest columns present: 1) the close-ups of the super-resolved images by different methods, 2) the epipolar plane images, and 3) the PSNR and SSIM. Best viewed zoom-in.}
    \label{fig:vis}
\end{figure*}

\begin{figure}[t]
    \centering
    \includegraphics[width=\linewidth]{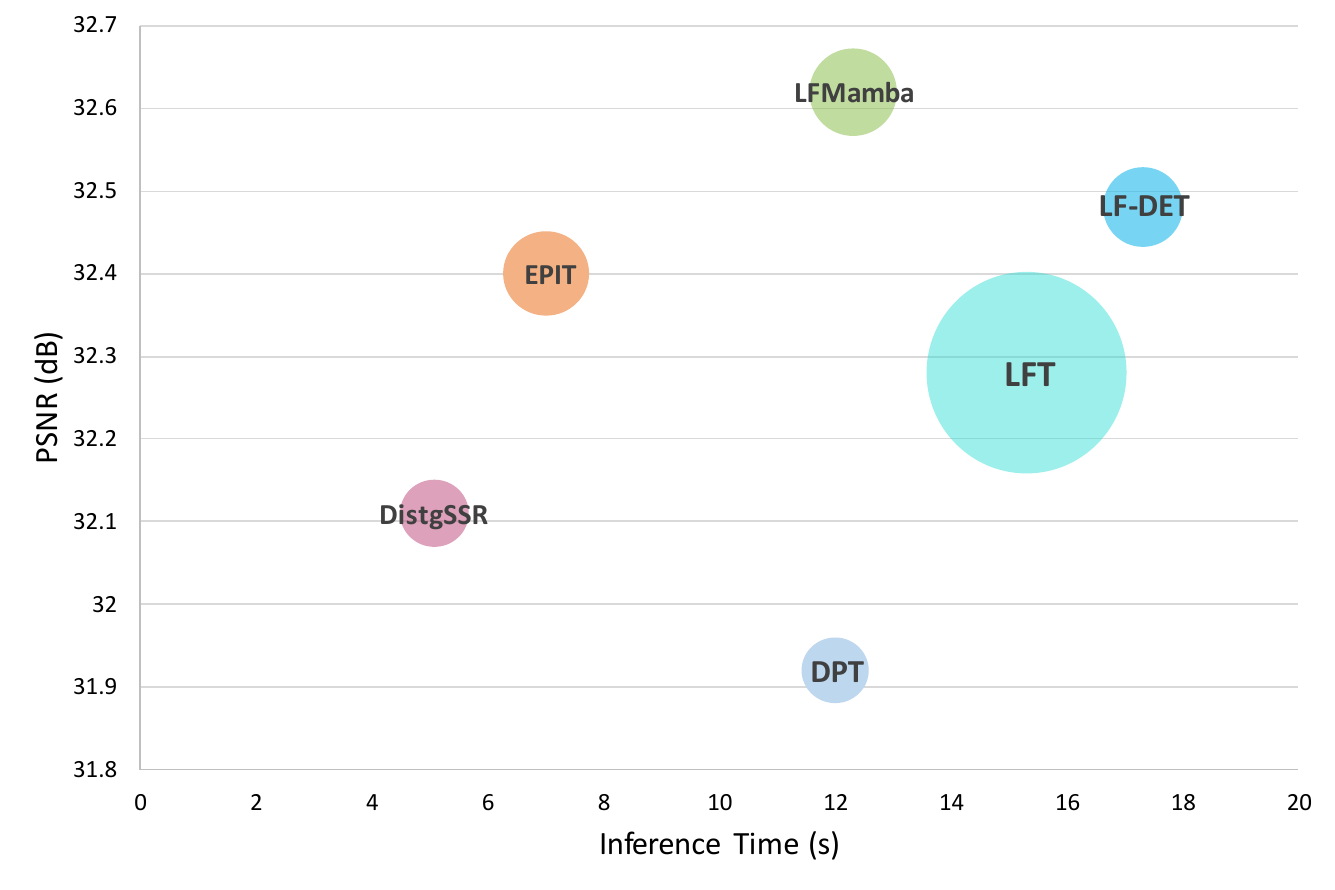}
    \caption{\textbf{Computational efficiency comparison between LFMamba and SOTA methods on $4\times$ SR.} The area of circles denotes the memory consumption. The inference time is calculated by averaging the inference time of all scenes across the five test datasets.  }
    \label{fig:eff}
\end{figure}

\subsection{Comparsion to state-of-the-art}
We compare our method with fifteen state-of-the-art methods, including one single image SR method \cite{rcan}, ten CNN-based LF image SR methods \cite{reslf, LFSSR, ATO, interNet, dfnet, megnet, iinet, distg, sav, hlfsr} and four Transformer-based LF image SR methods \cite{dpt, lft, epit, LF-DET}. We use PSNR and SSIM as quantitative metrics to evaluate the performance. For one dataset with $M$ scenes, we first get the average metric of $U\times V$ SAIs of each scene, then we obtain the metric of this dataset by averaging over $M$ scenes.

\noexpand{\textbf{Quantitative Results.}} Table \ref{tab:sota} presents the quantitative comparisons between our method and other state-of-the-art methods. LFMamba achieves competitive results across five datasets for both $\times2$ and $\times4$ SR while maintaining a moderate model size. Overall, LFMamba performs closely to LF-DET \cite{LF-DET} on $\times 2$ SR with only 0.03dB lower average PSNR but a 0.0007 higher average SSIM. While for the more challenging $\times 4$ SR, LFMamba surpasses LF-DET by an average of 0.14dB in PSNR and 0.0009 in SSIM. Particularly noteworthy are LFMamba's remarkable performance gains on real-world LF datasets EPFL and INRIA, which exhibit more intricate structures. This underscores LFMamba's capability to explore the intrinsic structure of LF effectively. However, it's also worth noting that LFMamba's performance on datasets with large disparities is somewhat unstable. For instance, while it outperforms LF-DET by 0.14dB in $\times 4$ SR on HCInew dataset (disparity range: [-4, 4]), it lags behind LF-DET and EPIT on the STFgantry dataset (disparity range: [-7, 7]) by 0.29dB and 0.33dB respectively. To address this imbalance, we employ a geometry assembling strategy to further enhance overall performance and narrow the performance gap on the STFgantry dataset between LFMamba, EPIT, and LF-DET, which brings an average PSNR increase of 0.17dB for $\times 2$ and 0.15dB for $\times 4$ SR, respectively. 

\noexpand{\textbf{Qualitative Results.}} We also present qualitative results achieved by different methods on more challenging $\times 4$ SR task. As depicted in Fig. \ref{fig:vis}, LFMamba excels in recovering sharp edges and intricate textures, a feat that eludes most competing methods. For instance, in the scene \textit{ISO\_Chart\_1} from the EPFL dataset, many methods struggle to reconstruct the horizontal lines between figures '2' and '3', whereas LFMamba accomplishes this task effectively. In the scene \textit{Red\_\&\_White\_Building} from the EPFL dataset, we can observe that only our LFMamba can reconstruct clear and complete diagonal red line of the no-parking marker. In the scene \textit{Origami} from HCnew dataset, LFMamab successfully recovers the spots and lines on the red jar, indicating the efficacy of LFMamba.

\noexpand{\textbf{Computational Efficiency.}} To evaluate the computational efficiency of LFMamba, we compare the model parameters, FLOPs, memory consumption, and inference time of different state-of-the-art models on $\times 4$ SR task. For parameters and FLOPs, we can see that from Table. \ref{tab:sota}, compared to the leading three CNN-based methods, i.e., IINet \cite{iinet}, DistgSSR \cite{distg}, and HLFSR \cite{hlfsr}, our LFMamba has fewer parameters or FLOPs while achieving better performance. Compared to Transformer-based methods, our LFMamba also keeps a moderate model size although LFMamba has more network layers, which is attributed to our proposed efficient S6 block.

Regarding inference time and memory consumption, we evaluate these metrics by averaging the inference time across 23 scenes from five test datasets and measuring peak memory usage on the GPU. Since convolution operation is inherently more efficient than self-attention calculation due to the computational complexity and modern GPU acceleration algorithms. We select one CNN-based method, DistgSSR \cite{distg} as the baseline model, and other Tansformer-based methods and our LFMamba for comparison. The results are shown in Fig. \ref{fig:eff}, our LFMamba obtains a good balance on performance and efficiency. In general, LFMamba's inference time is faster than LFT and LF-DET and is comparable to DPT. However, there is still a gap between LFMamba and EPIT as EPIT adopts self-attention on EPIs, which have a smaller feature size, and EPIT only has ten Transformer layers while our LFMamba has 24 SSM layers. Additionally, LFMamba consumes acceptable memory footprints (reflected by the circle size), indicating that it can effectively super-resolve LF images with limited resources. Note that, all model inferences are conducted in the same environment using an Nvidia 2080Ti GPU. 


\subsection{Ablation Study}
\noexpand{\textbf{LF Structure Feature Learning.}} As previously discussed, the structural characteristics of LFs is indispensable for reconstructing HR LF images. To probe the efficacy of our proposed LFMamba, we develop three model variants. \textbf{First}, we remove all the EPI-H/V SSM blocks and build a model with five spatial-angular SSM blocks, denoted as '\textit{w/o EPI}' to illustrate the necessity of LF structure learning. \textbf{Second}, we achieve LF structure learning through learning the relationships of bi-directional sub-aperture image sequences as introduced in Fig. \ref{fig:enter-label} \textbf{Mid}, denoted as '\textit{w/ 3D}'. \textbf{Last}, we invert the order of spatial-angular feature learning and LF structure learning, denoted as '\textit{w/ reverse}' to investigate the impact of the learning orders of spatial-angular information and LF structure. 

\begin{table*}[!]
    \centering
    \caption{\textbf{Ablation studies of LFMamba.} The \colorbox{gray!35}{first three rows} investigate the effect of LF structure learning. The \colorbox{gray!20}{second three rows} compares the effectiveness of Mamba and Transformer. The \colorbox{gray!10}{third two rows} investigate the multi-level feature fusion strategies.} 
    \resizebox{0.9\textwidth}{!}{
    \begin{tabular}{|c|c|cccccccccc|}
    \hline
         Model variants & \#Params.(M) & EPFL & $\Delta$ & HCInew & $\Delta$ & HCIold& $\Delta$ & INRIA& $\Delta$ & STFgantry& $\Delta$   \\
         \hline
         \textit{LFMamba} & 2.30 & 29.84 & - & 31.70 & - & 37.91 & - & 31.81 & - & 31.85 & -\\
         \hline
        \rowcolor{gray!35}\textit{w/o EPI} & 2.39 & 29.57 & {\color{red!90!black} -0.27} & 31.68& {\color{red!90!black} -0.02} & 37.79 & {\color{red!90!black} -0.12} & 31.46 & {\color{red!90!black} -0.35} & 32.01 & {\color{green!40!black} +0.16}\\
        \rowcolor{gray!35}\textit{w/ 3D} &  2.23 & 29.63 & {\color{red!90!black} -0.21} & 31.60 & {\color{red!90!black} -0.10} & 37.76 & {\color{red!90!black} -0.15} & 31.59 & {\color{red!90!black} -0.22}& 32.09 & {\color{green!40!black} +0.24}\\
        \rowcolor{gray!35}\textit{w/ reverse} & 2.30 & 29.58 & {\color{red!90!black} -0.26} & 31.58 & {\color{red!90!black} -0.12} & 37.74 & {\color{red!90!black} -0.17} & 31.45 & {\color{red!90!black} -0.36} & 31.75 & {\color{red!90!black} -0.10}\\
        \hline
        \rowcolor{gray!20}\textit{w/ LFT} & 1.65 & 29.70 & {\color{red!90!black} -0.14} & 31.66 &  {\color{red!90!black} -0.04}& 37.87 & {\color{red!90!black} -0.04} & 31.68 & {\color{red!90!black} -0.13} & 31.92 & {\color{green!40!black} +0.07}\\
        \rowcolor{gray!20}\textit{w/ EPIT} & 2.34  & 29.63 & {\color{red!90!black} -0.21} & 31.69 & {\color{red!90!black} -0.01}& 37.88 & {\color{red!90!black} -0.03} & 31.55 & {\color{red!90!black} -0.26} & 32.23 & {\color{green!40!black} +0.38}\\
        \rowcolor{gray!20}\textit{w/ LFT\_EPIT} & 2.26  & 29.59 & {\color{red!90!black} -0.25} & 31.62 & {\color{red!90!black} -0.08}& 37.94 & {\color{green!40!black} +0.03} & 31.57 & {\color{red!90!black} -0.24} & 32.21 & {\color{green!40!black} +0.36}\\
        \hline
        \rowcolor{gray!10}\textit{w/o fusion} & 2.17 & 29.63& {\color{red!90!black} -0.21} & 31.65 & {\color{red!90!black} -0.05}& 37.90 & {\color{red!90!black} -0.01}& 31.72 & {\color{red!90!black} -0.09} & 32.01 & {\color{green!40!black} +0.16} \\
        \rowcolor{gray!10}\textit{w sum} & 2.17 & 29.64& {\color{red!90!black} -0.20} & 31.70 & - & 37.85 & {\color{red!90!black} -0.05}& 31.65 & {\color{red!90!black} -0.16} & 31.92 & {\color{green!40!black} +0.07} \\
    \hline
    \end{tabular}
    }
    \label{tab:ablation}
\end{table*}

From the first three rows in Tabel \ref{tab:ablation}, we can conclude that. First, the absence of LF structure learning leads to a performance drop on most datasets, especially for real-world datasets (i.e., EPFL, INRIA). Second, learning LF structure information from 3D image sequence
perspective benefits the exploration of the STFgantry dataset, while is sub-optimal in other datasets. Because when the disparity is small, spatial and angular information contribute more to recovering fine-grained details. Third, extracting spatial-angular information is privileged, indicating that the effectiveness of LF structure learning can be maximized after spatial-angular feature learning. 

\begin{figure*}[t]
    \centering
    \includegraphics[width=0.9\linewidth]{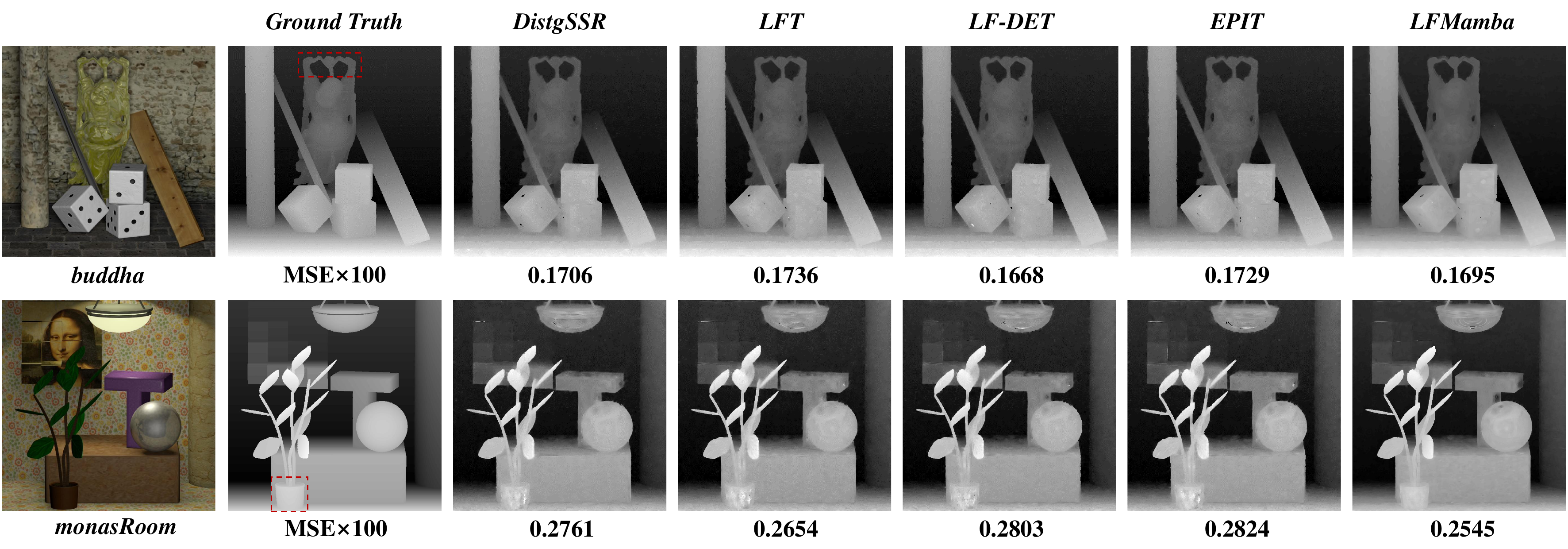}
    \caption{\textbf{Depth estimation results achieved by SPO \cite{SPO} using $\times 4$ SR LF images generated by different LFSR methods.} The mean square error multiplied by 100 (MSE $\times$ 100) is chosen as the quantitative metric.}
    \label{fig:depthl}
\end{figure*}

\noexpand{\textbf{Mamba \textit{vs.} Transformer.}} We develop three model variants '\textit{w/ LFT}', '\textit{w/ EPIT}', and '\textit{w/ LFT\_EPIT}' to investigate the performance comparison between Mamba and Transformer. Specifically, '\textit{w/ LFT}' means we replace the spatial and angular SSM with the spatial and angular Transformer in LFT \cite{lft}. '\textit{w/ EPIT}' means we substitute the EPI-V/H SSM with the non-local Transformer block proposed in EPIT \cite{epit}. And '\textit{w/ LFT\_EPIT}' denotes both happens. From the middle three rows in Table. \ref{tab:ablation}, we observe that, using Transforms on spatial and angular leads to an overall performance drop except for STFganrty with minor improvements. The integration of non-local Transformer block brings significant robustness to the STFgantry dataset, but also can not avoid performance drops on other datasets. 

\noexpand{\textbf{Multi-level Feature Fusion Strategies.}} We explore the impact of multi-level features (i.e., $\boldsymbol{F}_{init}$, $\boldsymbol{F}_{sa}$, and $\boldsymbol{F}_{struct}$) usage by introduce two model variants. \textbf{First}, we don't use feature fusion, denoted as '\textit{w/o fusion}'. This leads to an overall performance decline while surprisingly benefiting the STFgantry dataset, indicating that the utilization of multi-level features promotes the exploration of inherent LF information. \textbf{Second}, we replace the concatenation operation with the element summation operation, which we find even worse than the results of '\textit{w/o fusion}', demonstrating the effectiveness of our multi-level feature fusion strategy. 

\noexpand{\textbf{Numbers of the basic SSM block.}} We further explore the impact of the numbers of basic SSM block in spatial/angular/EPI-H/V SSM blocks. As shown in Table. \ref{tab:num}, on $\times 4$ SR task, when we utilize one basic SSM block, the model already reaches EPIT's \cite{epit} performance on average PSNR with the same parameters but lower FLOPs. However, employing three basic SSM blocks fails to bring consistent improvements. Therefore, we ultimately use two basic SSM blocks in LFMamba.

\begin{figure*}[ht]
    \centering
    \includegraphics[width=0.9\linewidth]{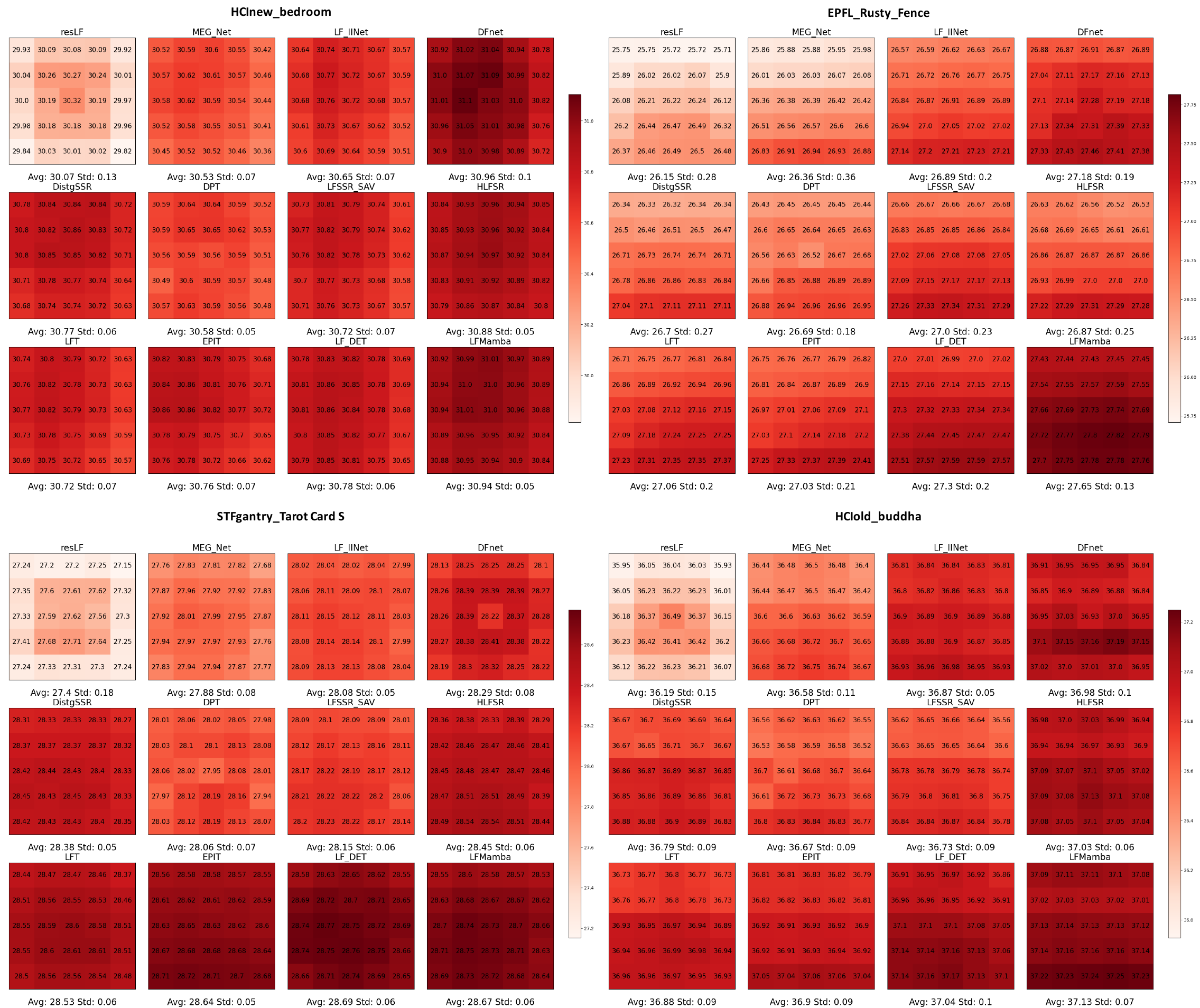}
    \caption{\textbf{Visulization of the PSNR distributions w.r.t. perspectives.} We also present the average PSNR and standard deviation for each method, lower standard deviation represents higher angular consistency.}
    \label{fig:perspective}
\end{figure*}
\begin{table}[!t]
    \centering
    \caption{Ablation studies of the numbers of basic SSM block on $\times 4$ SR.}
    \begin{tabular}{|c|ccc|}
    \hline
          Nums. & \#Params.(M) & FLOPs(G)&Average PSNR \\
         \hline
          1 & 1.47 & 43.17 & 32.40\\
          2 (LFMamba) & 2.30 &66.90& 32.62\\
          3 & 3.13 & 90.64 & 32.57\\
         \hline
    \end{tabular}
    \label{tab:num}
\end{table}

\subsection{Angular Consistency}
We also investigate the angular consistency of different models from the following three perspectives. \textbf{First}, we measure the angular consistency through the fidelity of the reconstructed EPIs. As shown in Fig. \ref{fig:vis}, for instance, in the scene \textit{ISO\_Chart\_1} from the EPFL dataset, the left part of the EPIs reconstructed by all the other methods present aliased results, while ours result is smoother which benefited from the LF structure learning. \textbf{Second}, we measure the angular consistency by using the $4 \times$ SR results for depth estimation using the SPO \cite{SPO} algorithm. We select \textit{buddha} and \textit{monasRoom} from HCIold datasets with ground truth depth map as test objectives and use mean square error as the quantitative metric. As shown in Fig. \ref{fig:depthl}, our LFMamaba achieves the best score on the scene \textit{monasRoom} and the second score on \textit{buddha}, which represents high angular consistency. \textbf{Last}, we evaluate the angular consistency by comparing the PSNR distributions of all views on some representative scenes for $4 \times$ SR. As shown in Fig. \ref{fig:perspective}, our LFMamba can achieve superior performance while obtaining a low standard deviation, showcasing that LFMamba can reconstruct high-quality HR LF images across all angular views, demonstrating its high angular consistency.

\begin{table}[]
    \centering
    \caption{\textbf{Quantitative comparison (PSNR/SSIM) between different methods in $2 \times 2$ to $7\times 7$ angular SR task.}The best results are \textbf{bolded}, and the second best are \underline{underlined}}.
    \resizebox{1\linewidth}{!}{
    \begin{tabular}{|c|c|ccc|}
    \hline
        Method & \#Params.(M) & HCIold & HCInew & Average\\
        \hline
         DistgASR \cite{distg} & 2.68 & \textbf{42.18/0.978} & 34.70/0.974 & 38.44/\textbf{0.976}\\
         EASR-L \cite{EASR}& 6.63 & 41.54/0.971& \textbf{35.86}/\underline{0.975} & \underline{38.70/0.973}\\
         LFMamba-ASR & 2.23& \underline{42.03/0.975} & \underline{35.61}/\textbf{0.977}& \textbf{38.82/0.976}\\
        \hline
    \end{tabular}}
    \label{tab:asr}
\end{table}

\subsection{LFMama for LF Angular SR}
To investigate the generalization ability of our proposed method, we apply our proposed LFMamba for LF angular SR (LFASR) task by simply modifying the HR LF Reconstruction module and remaining the rest part unchanged. Concretely, we take $2\times2 \xrightarrow{} 7\times7$ LFASR task as an example, after getting the deep fused feature (i.e., $\boldsymbol{F}_{fuse} \in \mathbb{R}^{2 \times 2 \times H \times W \times C}$), a $2\times 2$ convolution without padding is performed on the angular dimension the generate an angular sparse-downsampled feature $\boldsymbol{F}_{s} \in \mathbb{R}^{1 \times 1 \times H \times W \times C}$. Then, a $1\times 1$ convolution is applied to expand the channel followed by pixel-shuffle to generate angular dense-sampled feature $\boldsymbol{F}_{d} \in \mathbb{R}^{7 \times 7 \times H \times W \times C}$. Finally, a $3\times 3$ convolution is used to produce the final output $\mathcal{L} \in \mathbb{R}^{7\times 7 \times H \times W}$.

Following \cite{distg, EASR}, we select two datasets HCIold \cite{hciold} and HCInew \cite{hcinew} for experiments. We choose the central $7\times 7$ SAIs as high angular resolution LFs and crop them into $64\times64$ patches, and its corner $2\times 2$ SAIs as low angular resolution input LFs. During training LFMamba-ASR, we select $L_{1}$ loss and use Adam as the optimizer with $\beta_{1} = 0.9$ and $\beta_{2} = 0.999$. The initial learning rate is set to $2\times 10^{-4}$ and will be halved every 15 epochs in total 60 epochs and the batch size is set to 2. Data augmentation is also conducted as the same as LFMamba.  

\begin{figure}
    \centering
    \includegraphics[width=\linewidth]{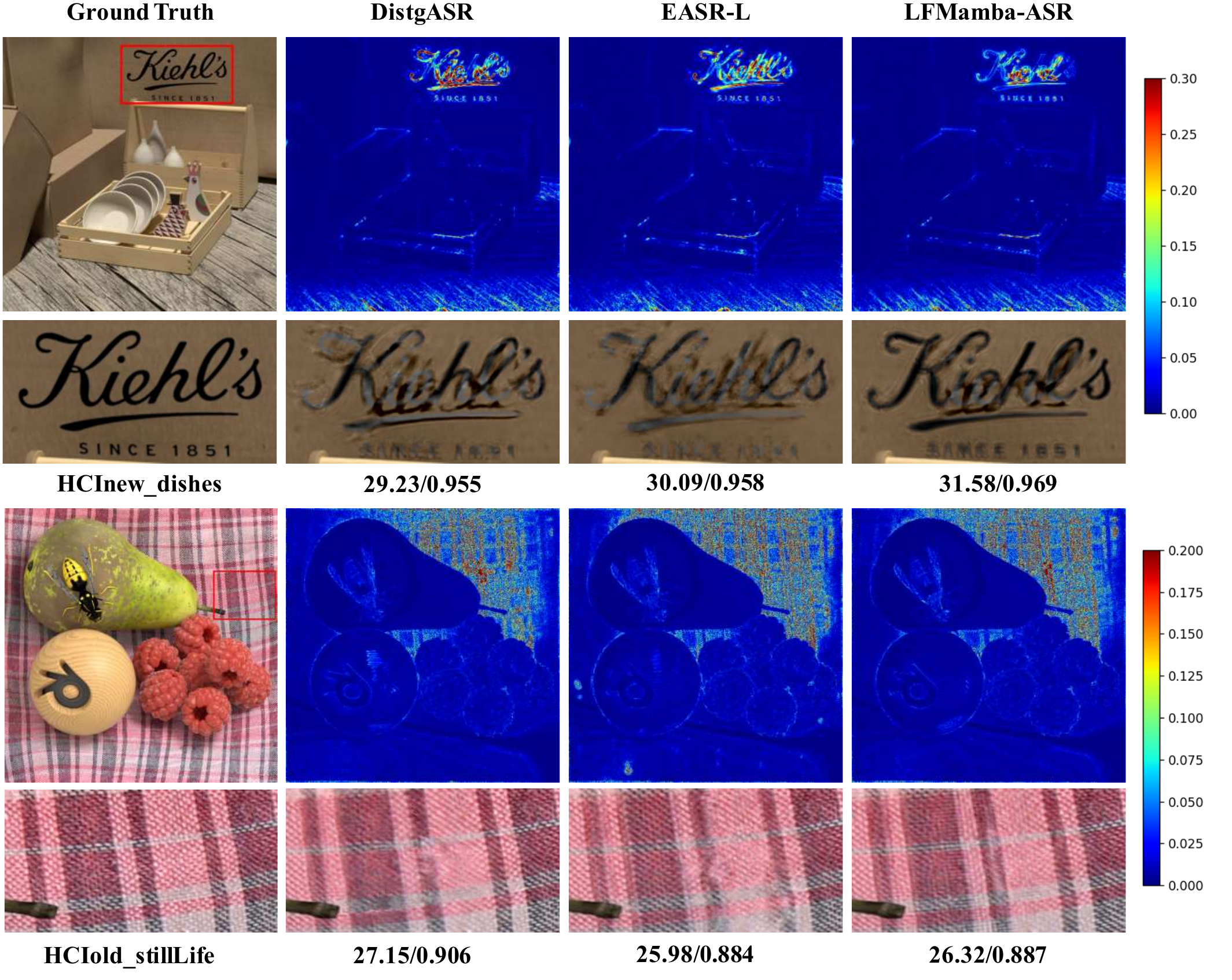}
    \caption{\textbf{Visual results of the reconstructed central view image achieved by different methods.} We also present the error map between the results and ground truth.}
    \label{fig:asr}
\end{figure}

We compare LFMamba-ASR with two state-of-the-art methods DistgASR \cite{distg} and EASR-L \cite{EASR}, both are CNN-based networks. The quantitative results are shown in Table. \ref{tab:asr}, although LFMamba-ASR ranks second place both in HCIold and HCInew dataset, it achieves the best average PSNR/SSIM score across these two datasets with the smallest model size. The qualitative results are shown in Fig. \ref{fig:asr}, we select two challenging scenes \textit{dishes} from HCInew and \textit{stillLife} from HCIold for comparisons. The reconstructed central view SAIs by LFMamba-ASR present fewer artifacts in the letter area of \textit{dishes} and fabric area of \textit{stillLife}. From the error maps, we can observe that LFMamba-ASR can reconstruct more accurate images. The promising results on LFASR demonstrate our proposed method's effectiveness and generalization abilities for utilizing SSM to learn LF features.


\section{Limitations and Future Work} 
Although LFMamba presents its competitive capability compared to existing leading LFSR methods, it still exhibits some limitations and can be further improved with more careful designs. For instance, as a pure SSM-based network, LFMamba is adept at recovering the sharp edges and fine-grained textures when the disparity is relatively small while encountering obstacles when the disparity is getting large. The experimental results indicate that a more robust retrofit on the basic SSM block or a combination of SSM and Transformer is worth trying. Additionally, since this work mainly focuses on the scanning approach to LFs, broader consideration of the exploration of other priors such as frequency analysis of LFs could be taken in future studies.

\section{Conclusion}
In this work, we integrate the recent advanced Selective State Space Model, i.e., Mamba into Light Field image Super-Resolution (LFSR). We start with analyzing the probable approaches and ultimately choosing to employ Mamba on LFs' informative 2D slices to comprehensively explore the spatial contextual information, complementary angular information, and LF structure information. Based on that, we propose LFMamba, a pure SSM-based network built upon our proposed basic SSM block that is characterized by the proposed efficient SS2D mechanism, which facilitates efficient and effective exploration of the rich LF information. Extensive experiments demonstrate the competitiveness in terms of performance and efficiency of LFMamba against state-of-the-art CNN- and Transformer-based LFSR methods. We further explore the generalization ability of our method by applying LFMamba for LFASR task and achieving promising results, further demonstrating that our proposed method can effectively learn inherent LF features from a representation learning perspective.

\small{
\bibliographystyle{IEEEtranN}
\bibliography{mybib}}

\end{document}